\PassOptionsToPackage{dvipsnames}{xcolor}
\documentclass[runningheads]{llncs}
\usepackage{graphicx}
\usepackage{tikz}
\usepackage{comment}
\usepackage{amsmath,amssymb} %
\usepackage{bbm}
\usepackage{color}
\usepackage{tabularx}
\usepackage{etoolbox}
\usepackage{booktabs}
\usepackage{arydshln}
\usepackage{makecell}
\usepackage{colortbl}
\usepackage{tabu}
\usepackage{multirow}
\usepackage{enumitem}
\usepackage{setspace}
\usepackage{breakcites}

\definecolor{bblue}{rgb}{0.0,0.25,0.75}
\definecolor{ccol}{rgb}{0.2,0.2,0.2}

\usepackage[pagebackref=true,breaklinks=true,colorlinks=true,urlcolor=bblue,bookmarks=false,citecolor=bblue]{hyperref}

\usepackage[capitalise]{cleveref}

\usepackage[skip=1ex,font=small,labelsep=period]{caption}
\usepackage[accsupp]{axessibility}  %

\newcolumntype{Y}{>{\centering\arraybackslash}X}
\newcolumntype{R}{>{\raggedleft\arraybackslash}X}
\newcolumntype{L}{>{\raggedright\arraybackslash}X}

\newcommand\customparagraph[1]{\vspace{0.7em}\noindent\textbf{#1}}
\renewcommand{\mathbbm}[1]{\text{\usefont{U}{bbm}{m}{n}#1}}

\newcommand{\wincat}[1]{\textcolor{black}{\textbf{#1}}}

\usepackage{xspace}
\newcommand*{\eg}{\emph{e.g.}\@\xspace}
\newcommand*{\ie}{\emph{i.e.}\@\xspace}
\newcommand*{\wrt}{\emph{w.r.t.}\@\xspace}

\definecolor{lightgray}{rgb}{0.835, 0.835, 0.835}
\definecolor{lightergray}{rgb}{0.935, 0.935, 0.935}

\makeatletter
\def\adl@drawiv#1#2#3{%
        \hskip.5\tabcolsep
        \xleaders#3{#2.5\@tempdimb #1{1}#2.5\@tempdimb}%
                #2\z@ plus1fil minus1fil\relax
        \hskip.5\tabcolsep}
\newcommand{\cdashlinelr}[1]{%
  \noalign{\vskip\aboverulesep
           \global\let\@dashdrawstore\adl@draw
           \global\let\adl@draw\adl@drawiv}
  \cdashline{#1}
  \noalign{\global\let\adl@draw\@dashdrawstore
           \vskip\belowrulesep}}
\makeatother

\makeatletter
\newlength{\qrr@dimen@}
\expandafter\pretocmd\csname tabular*\endcsname{\setlength{\qrr@dimen@}{#1}}{}{}
\newcommand*{\Rowcolor}[2][\tabcolsep]{%
    \ifx\relax#1\relax\else
        \kern-\the\dimexpr#1\relax
    \fi
    \makebox[0pt][l]{%
        \fboxsep=0pt
        \colorbox{#2}{%
            \strut\kern\qrr@dimen@
        }%
    }%
    \ifx\relax#1\relax\else
        \kern\the\dimexpr#1\relax
    \fi
    \ignorespaces
}
\makeatother

\begin{document}

\pagestyle{headings}
\mainmatter
\def\ECCVSubNumber{7039}  %

\title{Neural Correspondence Field\\for Object Pose Estimation}

\titlerunning{Neural Correspondence Field for Object Pose Estimation}

\newcommand{\namesep}{\hspace{0.8em}}
\author{
    Lin Huang{\small $^{1*}$}\namesep
    Tomas Hodan{\small $^{2}$}\namesep
    Lingni Ma{\small $^{2}$}\namesep
    Linguang Zhang{\small $^{2}$}\\
    Luan Tran{\small $^{2}$}\namesep
    Christopher Twigg{\small $^{2}$}\namesep
    Po-Chen Wu{\small $^{2}$}\\
    Junsong Yuan{\small $^{1}$}\namesep
    Cem Keskin{\small $^{2}$}\namesep
    Robert Wang{\small $^{2}$}
}
\institute{
 {$^{1}$University at Buffalo}\namesep
 {$^{2}$Reality Labs at Meta}
}

\authorrunning{Huang, Hodan, Ma, Zhang, Tran, Twigg, Wu, Yuan, Keskin, Wang}
\maketitle

\begin{abstract}
We propose a method for estimating the 6DoF pose of a~rigid object with an available 3D model from a single RGB image.
Unlike classical correspondence-based methods which predict 3D object coordinates at pixels of the input image, the proposed method predicts 3D object coordinates at 3D query points 
sampled in the camera frustum.
The move from pixels to 3D points, which is inspired by recent PIFu-style methods for 3D reconstruction, enables reasoning about the whole object, including its (self-)occluded parts.
For a 3D query point associated with a~pixel-aligned image feature, we train a fully-connected neural network to predict: (i)~the corresponding 3D object coordinates, and (ii)~the signed distance to the object surface, with the first defined only for query points in the surface vicinity.
We call the mapping realized by this network as {\it Neural Correspondence Field}.
The object pose is then robustly estimated from the predicted 3D-3D correspondences by the Kabsch-RANSAC algorithm.
The proposed method achieves state-of-the-art results on three BOP datasets and is shown superior especially in challenging cases with occlusion. The project website is at: \texttt{\href{https://linhuang17.github.io/NCF}{linhuang17.github.io/NCF}}.

\end{abstract}

\section{Introduction}

Estimating the 6DoF pose of a rigid object is a fundamental computer vision problem with great importance to application fields such as augmented reality and robotic manipulation.
In recent years, the problem has received considerable attention and the state of the art has improved substantially, yet there remain challenges to address, particularly around robustness to object occlusion~\cite{hodan2018bop,hodan2020bop}.

{\let\thefootnote\relax\footnotetext{$^{*}$Work done during Lin Huang's internship with Reality Labs at Meta.}}

Recent PIFu-style methods for 3D reconstruction from an RGB image \cite{saito2019pifu,saito2020pifuhd,huang2020arch,zheng2021pamir,kulkarni2021s} rely on 3D implicit representations and demonstrate the ability to learn and incorporate strong priors about the invisible scene parts. For example, PIFu~\cite{saito2019pifu} is able to faithfully reconstruct a 3D model of the whole human body, and DRDF~\cite{kulkarni2021s} is able to reconstruct a 3D model of the whole indoor scene, including parts hidden behind a couch.
Inspired by these results, we propose a 6DoF object pose estimation method based on a 3D implicit representation and analyze its performance specifically in challenging cases with occlusion.

\begin{figure}[t!]
	\begin{center}
	    \setlength{\tabcolsep}{6pt}
        \begin{tabularx}{\linewidth}{Y Y Y Y}
            {\small RGB input\phantom{----}} & \multicolumn{2}{c}{{\small Predicted 3D-3D correspondences}} & {\small \phantom{------}Estimated pose}
            \vspace{0.5ex}
        \end{tabularx}
	    \includegraphics[width=1.0\linewidth]{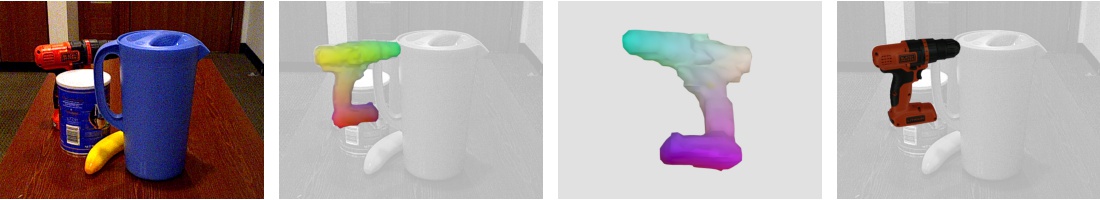}\vspace{0.7ex}
		\includegraphics[width=1.0\linewidth]{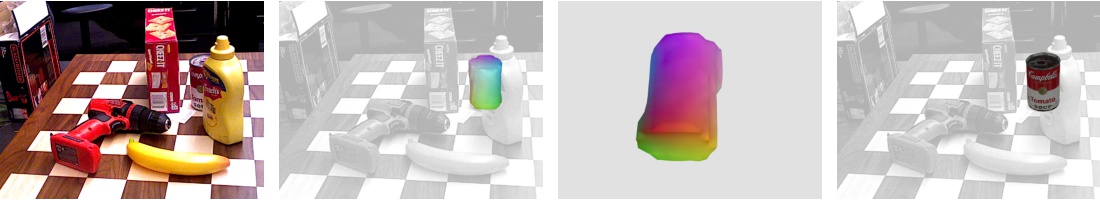}\vspace{0.7ex}
		\includegraphics[width=1.0\linewidth]{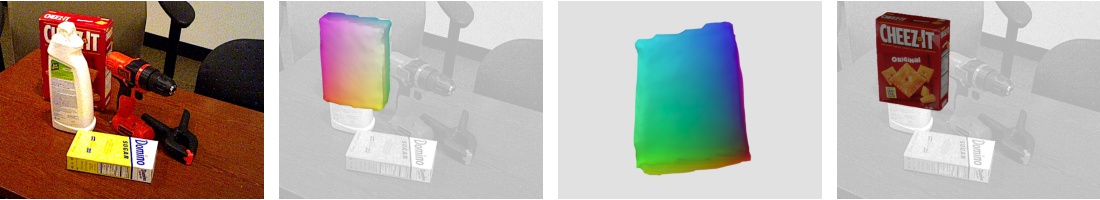}\vspace{0.7ex}
        \caption{\label{fig:teaser}
\textbf{An overview of the proposed method.} The object pose is estimated from 3D-3D correspondences established by predicting 3D object coordinates at 3D query points densely sampled in the camera frustum. For efficient selection of reliable correspondences nearby the object surface, the method predicts for each query point also the signed distance to the surface. The middle columns show two views of a mesh that is reconstructed by Marching Cubes~\cite{lorensen1987marching} from the predicted signed distances and colored with the predicted 3D object coordinates (the mesh is reconstructed only
for visualization purposes, not when estimating the object pose). The 3D CAD model, which
is assumed available for each object, is shown in the estimated pose on the right.
}
	\end{center}
\end{figure}

Similarly to PIFu~\cite{saito2019pifu}, the proposed method makes predictions for 3D query points that are sampled in the camera frustum and associated with pixel-aligned image features. 
PIFu predicts color and occupancy, \ie, a binary signal that indicates whether a query point is inside or outside the object. Instead, the proposed method predicts (i)~the corresponding 3D object coordinates, and (ii)~the signed distance to the object surface, with the first defined only for query points in the surface vicinity, \ie, points for which the predicted signed distance is below a threshold.
The 6DoF object pose is then robustly estimated from the predicted 3D-3D correspondences between 3D query points and the predicted 3D object coordinates by the Kabsch algorithm~\cite{kabsch} in combination with RANSAC~\cite{fischler1981random}.

Classical methods for 6DoF object pose estimation~\cite{collet2011moped,brachmann2014learning,park2019pix2pose,zakharov2019dpod,rad2017bb8,tekin2018real,peng2019pvnet,hodan2020epos} rely on 2D-3D correspondences established between pixels of the input image and the 3D object model, and estimate the pose by the P\emph{n}P-RANSAC algorithm~\cite{lepetit2009epnp}. The proposed method predicts 3D object coordinates for 3D query points instead of pixels. This enables reasoning about the whole object surface, including self-occluded parts and parts occluded by other objects.
In Sec.~\ref{sec:experiments}, we show that the proposed method noticeably outperforms a baseline method that relies on the classical 2D-3D correspondences. Besides, we show that the proposed method outperforms all existing methods with the same training and evaluation setup (\ie, RGB-only and without any iterative refinement of pose estimates) on datasets YCB-V, LM-O, and LM from the BOP benchmark~\cite{hodan2018bop,hodan2020bop}.

\vspace{2.0ex}
\noindent
This work makes the following contributions:
\vspace{-2.0ex}

\begin{enumerate} %
\item The first method for 6DoF object pose estimation which demonstrates the effectiveness of a 3D implicit representation in solving this problem.
\item Neural Correspondence Field (NCF), a learned 3D implicit representation defined by a mapping from the camera space to the object model space, is used to establish 3D-3D correspondences from a single RGB image.
\item The proposed method noticeably outperforms a baseline based on 2D-3D correspondences
and achieves state-of-the-art results on three BOP datasets.
\end{enumerate}

\section{Related Work}

\noindent\textbf{6DoF Object Pose Estimation.}
Early methods for 6DoF object pose estimation assumed a grayscale or RGB input image and relied on local image features~\cite{lowe1999object,collet2011moped} or template matching~\cite{brunelli2009template}. 
After the introduction of Kinect-like sensors, methods based on RGB-D template matching~\cite{hinterstoisser2012accv,hodan2015detection}, point-pair features~\cite{drost2010model,hinterstoisser2016going,vidal2018method}, 3D local features~\cite{guo2016comprehensive}, and learning-based methods~\cite{brachmann2014learning,tejani2014latent,krull2015learning} demonstrated superior performance over RGB-only counterparts. Recent methods are based on convolutional neural networks (CNNs) and focus primarily on estimating the pose from RGB images. In the 2020 edition of the BOP challenge~\cite{hodan2020bop}, CNN-based methods finally caught up with
methods based on point-pair features which were dominating previous editions of the challenge. A popular approach adopted by the CNN-based methods is to establish 2D-3D correspondences by predicting 3D object coordinates at densely sampled pixels, and robustly estimate the object pose by the P\emph{n}P-RANSAC algorithm~\cite{ipose:jafari:accv18, nocs:wang:cvpr19, zakharov2019dpod, park2019pix2pose, li2019cdpn, hodan2020epos}. In Sec.~\ref{sec:experiments}, we show that our proposed method outperforms a baseline method that follows the 2D-3D correspondence approach and shares implementation of the common parts with the proposed method.
Methods establishing the correspondences in the opposite direction, \ie, by predicting the 2D projections of a fixed set of 3D keypoints pre-selected for each object model, have also been proposed~\cite{rad2017bb8,pavlakos20176,oberweger2018making,tekin2018real,tremblay2018deep,hu2019segmentation,peng2019pvnet}.
Other approaches localize the objects with 2D bounding boxes, and for each box predict the pose by regression~\cite{xiang2017posecnn,li2018unified,manhardt2019explaining,labbe2020cosypose} or classification into discrete viewpoints~\cite{kehl2017ssd,corona2018pose,sundermeyer2019augmented}.
However, in the case of occlusion, estimating accurate 2D bounding boxes covering the whole object, including the invisible parts, is problematic~\cite{kehl2017ssd}.

\customparagraph{Shape Reconstruction with Implicit Representations.}
Recent works have shown that a 3D shape can be modeled by a continuous and differentiable implicit representation realized by a fully-connected neural network. Examples of such representations include signed distance fields (SDF)~\cite{deepsdf2019park, igr:gropp:icml20, sal:atzmon:cvpr20, sald:atzmon:iclr21, metasdf:sitzmann:2019}, which map a 3D query point to the signed distance from the surface, and binary occupancy fields~\cite{occnet:mescheder:cvpr19,lif:chen:cvpr19,liu2021fully}, which map a 3D query point to the occupancy value.
Following the success of implicit representations, GraspingField~\cite{graspfield:3dv2020} extends the idea to reconstructing hands grasping objects. Instead of learning a single SDF, the method learns one SDF for hand and one for object, which allows to directly enforce physical constraints such as no interpenetration and proper contact.

For image-based reconstruction, Texture fields~\cite{texfield:oechsle:iccv19} learn textured 3D models by mapping a shape feature, an image feature, and a 3D point to color. OccNet~\cite{occnet:mescheder:cvpr19} proposes to condition occupancy prediction on an image feature extracted by a CNN. DISN~\cite{disn:xu:neurips19} improves this technique by combining local patch features with a global image feature to estimate SDF for 3D query points. PIFu~\cite{saito2019pifu}, which is closely related to our work, first extracts an image feature map by an hourglass CNN and then applies a fully-connected neural network to map a pixel-aligned feature with the depth of a 3D query point to occupancy. 
The follow-up work, PIFuHD~\cite{saito2020pifuhd}, recovers more detailed geometry by leveraging the surface normal map and multi-resolution volumes. PIFu and PIFuHD focus on human digitization. As for many other methods, experiments are done on images with cleanly segmented foreground. Recently, NeRF-like
methods reported impressive results in scene modeling~\cite{snr:sitzmann:neurips19, nv:lombardi:tog19, dvr:niemeyer:cvpr20, idr:yariv:neurips20, nerf:eccv20}. 
These methods typically require multi-view images with known camera calibration. For an in-depth discussion, we refer to the survey in~\cite{nrservey21}. In this work, we focus on a single input image and reconstruct known objects in unknown poses that we aim to recover.

\customparagraph{Learning Dense Correspondences.}
One of the pioneering works that learns dense correspondences is proposed in~\cite{scr:shotton:cvpr13} for camera relocalization, and extended for pose estimation of 
specific rigid objects in~\cite{brachmann2014learning, brachmann2016uncertainty, michel2017global}. These methods predict 3D scene/object coordinates at each pixel of the input image by a random forest. Later methods predict the coordinates by a CNN~\cite{ipose:jafari:accv18, park2019pix2pose, li2019cdpn, zakharov2019dpod, hodan2020epos}.
NOCS~\cite{nocs:wang:cvpr19} defines normalized object coordinates for category-level object pose estimation. Besides correspondences for object pose estimation, DensePose~\cite{densepose:guler:cvpr18} densely regresses part-specific UV coordinates for human pose estimation. CSE~\cite{cse:neverova:cvpr20} extends the idea to predict correspondences for deformable object categories by regressing Laplace-Beltrami basis and is extended to model articulated shapes in~\cite{articulatecsm:kulkarni:cvpr20}. These methods focus on learning mapping from pixels to 3D coordinates.
DIF-Net~\cite{dif:deng:cvpr21} jointly learns the shape embedding of an object category and 3D-3D correspondences with respect to a template. Similarly, NPMs~\cite{npms:pablo:iccv21} learns a 3D 
deformation field to model deformable shapes. Recent methods~\cite{dnerf:pumarola:cvpr20, nerfies:park:iccv21,nrnerf:edgar:iccv21} model deformable shapes by learning radiance and deformation fields.
None of these methods aims to recover the pose from images.

\section{Preliminaries}\label{sec:prelimilaries}
\subsubsection{Notations.} An RGB image is denoted by $I:\mathbb{R}^2\mapsto\mathbb{R}^3$
and can be mapped to a feature map $F:\mathbb{R}^2\mapsto\mathbb{R}^{K}$ with $K$ channels by an hourglass neural network~\cite{newell2016stacked,saito2019pifu}.
A 3D point $\mathbf{x}=[x,y,z]^\top\in\mathbb{R}^3$ in the camera coordinate frame can be projected to a pixel $[u,v]^\top\in\mathbb{R}^2$ by the projection function $\pi(\mathbf{\mathbf{x}}):\mathbb{R}^3\mapsto\mathbb{R}^2$.
Without loss of generality, we use a pinhole camera model with the projection function defined as: $\pi(\mathbf{x}) = \left[xf_x/z + c_x, \, yf_y/z+c_y\right]^\top$, where $f_x, f_y$ is the focal length and $(c_x, c_y)$ is the principal point.

A 6DoF object pose is defined as a rigid transformation $(R, \mathbf{t})$,
where $R\in\mathbb{SO}(3)$ is a 3D rotation matrix and $\mathbf{t}\in\mathbb{R}^3$ is a 3D translation vector. A 3D point~$\mathbf{y}$ in the model coordinate frame (also referred to as \emph{3D object coordinates}~\cite{brachmann2014learning}) is transformed to a 3D point~$\mathbf{x}$ in the camera coordinate frame as: $\mathbf{x} = R \mathbf{y} + \mathbf{t}$.

\customparagraph{Signed Distance Function (SDF)~\cite{tsdf:curless96, deepsdf2019park}.} In the proposed method, the object surface is represented implicitly with a
signed distance function, $\psi(\mathbf{x}): \mathbb{R}^{3}\mapsto\mathbb{R}$, which maps a 3D point $\mathbf{x}$ to the signed distance between $\mathbf{x}$ and the object surface. The signed distance is zero on the object surface, positive if $\mathbf{x}$ is outside the object and negative if $\mathbf{x}$ is inside.

\customparagraph{Kabsch Algorithm~\cite{kabsch}.} Given $N \geq 3$ pairs of corresponding 3D points $X=\{\mathbf{x}_i\}_N$ and $Y=\{\mathbf{y}_i\}_N$, the Kabsch algorithm finds a rigid transformation that aligns the corresponding 3D points by minimizing the following least squares:
\begin{equation}
    R^\star,\mathbf{t}^\star = \underset{R, \mathbf{t}}{\arg\min} 
    \sum_i^N \left\| R\mathbf{y}_i + \mathbf{t} - \mathbf{x}_i\right\|_2.
\end{equation}
The 3D rotation is solved via SVD of the covariance matrix: $USV^\top=\text{Cov}(X-\mathbf{c}_X,Y-\mathbf{c}_Y)$, $R^\star=VU^\top$, where $\mathbf{c}_X$ and $\mathbf{c}_Y$ are centroids of the point sets $X$ and $Y$ respectively. To ensure right-handed coordinate system, the signs of the last column of matrix $V$ are flipped if $\det(R^\star)=-1$~\cite{kabsch}. The 3D translation is then calculated as: $\mathbf{t}^\star = \mathbf{c}_X - R^\star\mathbf{c}_Y$. In the proposed method, we combine the Kabsch algorithm with a RANSAC-style fitting scheme~\cite{fischler1981random} to estimate the object pose from 3D-3D correspondences.

\customparagraph{PIFu~\cite{saito2019pifu}.} The PIFu method reconstructs 3D models of humans from segmented single/multi-view RGB images. For the single-view inference, the method first obtains a feature map $F$ with an hourglass neural network. Then it applies a fully-connected neural network, $f_\text{PIFu}(F(\pi(\mathbf{x})), \mathbf{x}_z) = o$, to map a pixel-aligned feature $F(\pi(\mathbf{x}))$ and the depth $\mathbf{x}_z$ of a 3D query point $\mathbf{x}$ to the occupancy $o\in[0,1]$ ($1$ means the 3D point is inside the model and $0$ means it is outside).

\section{The Proposed Method}\label{sec:method}

This section describes the proposed method for estimating the 6DoF object pose from an RGB image.
The image is assumed to show a single target object,
potentially with clutter, occlusion, and diverse lighting and background. In addition, the 3D object model, camera intrinsic parameters, and a large set of training images annotated with ground-truth object poses are assumed available.

The proposed method consists of two stages: (1) prediction of 3D-3D correspondences between the camera coordinate frame and the model coordinate frame (Sec.~\ref{sec:ncf}), and (2) fitting the 6DoF object pose to the predicted correspondences using the Kabsch-RANSAC algorithm (Sec.~\ref{sec:kabsch}).

\begin{figure}[b!]
    \vspace{-1.0ex}
    \centering
	\includegraphics[width=1.0\linewidth]{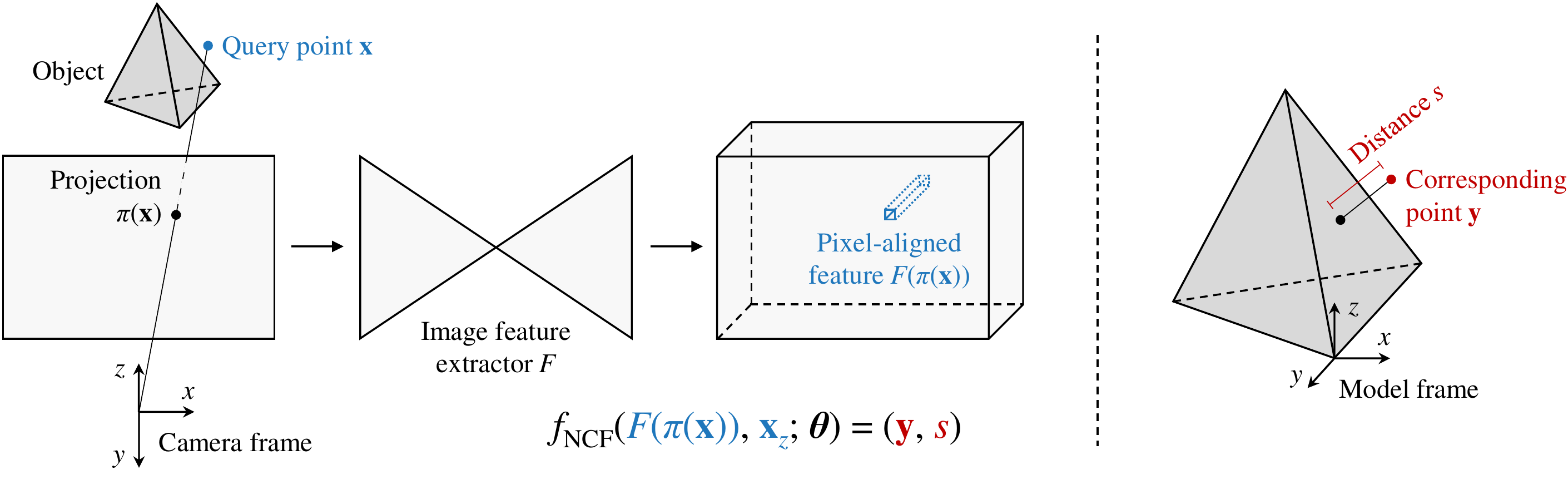}
	\caption{\textbf{Neural Correspondence Field} is a mapping learned by a fully-connected neural network $f_\text{NCF}$ with parameters $\boldsymbol{\theta}$. The input of the network is (i)~an image feature $F(\pi(\mathbf{x}))$ extracted at the 2D projection $\pi(\mathbf{x})$ of a 3D query point $\mathbf{x}$ sampled in the camera frustum, and (ii)~the depth $\mathbf{x}_z$ of $\mathbf{x}$. The output is (i)~the corresponding 3D point $\mathbf{y}$ in the model frame, and (ii)~the signed distance $s$ between $\mathbf{y}$ and the object surface. The point $\mathbf{y}$ is defined only if $|s|$ is below a fixed clamping threshold~$\delta$.}
	\label{fig:ncf}
\end{figure}

\subsection{Predicting Dense 3D-3D Correspondences}\label{sec:ncf}
\subsubsection{Neural Correspondence Field (NCF).} The 3D-3D correspondences are established using NCF defined as a mapping from the pixel-aligned feature $F(\pi(\mathbf{x}))$ and the depth $\mathbf{x}_z$ of a 3D query point $\mathbf{x}$ in the camera frame to the corresponding 3D point~$\mathbf{y}$ in the model frame and its signed distance $s$
(see also Fig.~\ref{fig:ncf}):
\begin{equation}\label{eq:ncf}
    f_\text{NCF}: \mathbb{R}^K \times \mathbb{R} \mapsto \mathbb{R}^3 \times \mathbb{R}     \text{~~as~~}
    f_\text{NCF}\big(F(\pi(\mathbf{x})), \mathbf{x}_z; \boldsymbol{\theta}\big) = \big(\mathbf{y}, s\big),
\end{equation}
where $\boldsymbol{\theta}$ are parameters of a fully-connected neural network $f_\text{NCF}$ that realizes the mapping. %
In our experiments, $f_\text{NCF}$
has the same architecture as the fully-connected network in PIFu~\cite{saito2019pifu}, except the output dimension is $4$ and $\text{tanh}$ is used as an activation function in the last layer, as in~\cite{deepsdf2019park}.
The feature extractor $F$ is realized by the hourglass neural network from PIFu and is applied to the input image remapped to a reference pinhole camera (arbitrarily chosen). The remapping is important to keep the depth $\mathbf{x}_z$ in accord with the image feature $F(\pi(\mathbf{x}))$ across images captured by cameras with different focal lengths.

Compared to PIFu, NCF additionally predicts the corresponding 3D point~$\mathbf{y}$, which enables establishing 3D-3D correspondences that are used for object pose fitting.
Besides, NCF predicts the signed distance
instead of the binary occupancy. This enables efficient selection of near-surface correspondences by thresholding the signed distances. Using near-surface correspondences increases pose fitting accuracy as learning correspondences from images with diverse background becomes ill-posed for 3D points far from the surface.
Since the 3D object model is available, the signed distance $s$ could be calculated from the predicted 3D point $\mathbf{y}$. However, we chose to predict the signed distance explicitly to speed up the method at both training and test time (predicting the value explicitly takes virtually no extra time nor resources).

\customparagraph{Training.} With parameters of the hourglass network $F$ denoted as $\boldsymbol{\eta}$ and parameters of the NCF network $f_{\text{NCF}}$ denoted as $\boldsymbol{\theta}$, the two networks are trained jointly by solving the following optimization problem:
\begin{align}\label{eq:loss}
    \boldsymbol{\eta}^\star, \boldsymbol{\theta}^\star = \underset{\boldsymbol{\eta,\theta}}{\arg\min} \;
    L_\mathbf{y} + \lambda L_{s},
\end{align}
where $L_\mathbf{y}$ and $L_s$ are regression losses on the 3D point $\mathbf{y}$ and the signed distance $s$, respectively. The scalar $\lambda$ is a balancing weight.
Assuming $N$ 3D points sampled in the camera frustum per image, the losses are defined as:
\begin{align}
    L_\mathbf{y} &= \min_{(\bar{R}, \bar{\mathbf{t}}) \in S} 
    \frac{1}{N} \sum_{i} 
    \mathbbm{1}\left(\left|\psi\left(\bar{\mathbf{y}}_i\right)\right| < \delta\right)
    H\left(\bar{R}\mathbf{y}_i + \bar{\mathbf{t}}, \mathbf{x}_i\right), \\
    L_s &= \frac{1}{N} \sum_{i}
    \big| \text{clamp}\left(\psi(\bar{\mathbf{y}}_i), \delta\right) - \text{clamp}\left(s, \delta\right) \big|,
    \label{eq:loss_y}
\end{align}
where $(\bar{R}, \bar{\mathbf{t}})$ is a ground-truth pose, $\bar{\mathbf{y}}_i = \bar{R}^{-1}(\mathbf{x}_i - \bar{\mathbf{t}})$ is the ground-truth corresponding 3D point, and $\delta$ is a clamping parameter controlling the distance from the surface over
which we expect to maintain a metric SDF, as in~\cite{deepsdf2019park}. The indicator function $\mathbbm{1}(\cdot)$ selects points within the clamping distance, and $H$
is the Huber loss~\cite{huber1992robust}.
To handle symmetric objects, we adopt the approach from NOCS~\cite{nocs:wang:cvpr19} which uses a pre-defined set of symmetry transformations (continuous symmetries are discretized) to get a set $S$ of possible ground-truth poses.

\customparagraph{Sampling 3D Query Points.} 
Given a training image
and the ground-truth object pose, the 3D object model is first transformed to the camera frame to assist with sampling of the query points.
As the training images may show the object in diverse scenes,
we found it crucial to focus the training on the object by sampling the query points more densely around the object surface.
In our experiments, we first sample three types of points: 12500 points nearby the surface, 1000 points inside the bounding sphere of the model, and 1000 points inside the camera frustum.
From these points, we sample 2500 points inside the model and 2500 points outside.
Note that this sampling strategy is invariant to occlusion, which forces the network $f_{\text{NCF}}$ to learn the complete object surface.
At test time, with no knowledge of the object pose, the points are sampled at centers of voxels that fill up the camera frustum in a specified depth range.

\subsection{Pose Fitting}\label{sec:kabsch}

To estimate the 6DoF object pose at test time, a set of 3D-3D correspondences, $C=\{(\mathbf{x}_i, \mathbf{y}_i)\}_M$ with $M \geq 3$, is established by linking each 3D query point $\mathbf{x}$ with the predicted 3D point $\mathbf{y}$ for which the predicted signed distance $s$ is below the threshold~$\delta$.
The object pose is then estimated from $C$ by a RANSAC-style fitting scheme~\cite{fischler1981random}, which iteratively proposes a pose hypothesis by sampling a random triplet of 3D-3D correspondences from $C$ and calculating the pose from the triplet by the Kabsch algorithm detailed in Sec.~\ref{sec:prelimilaries}. The quality of a pose hypothesis $(R,\mathbf{t})$ is measured by the number of inliers, \ie, the number of correspondences $(\mathbf{x}, \mathbf{y}) \in C$ for which $\|R \mathbf{y} + \mathbf{t} - \mathbf{x} \|_2$ is below a fixed threshold~$\tau$. In the presented experiments, a~fixed number of pose hypotheses is generated for each test image, and the final pose estimate is given by the hypothesis of the highest quality which is further refined by the Kabsch algorithm applied to all inliers. Note that the pose is not estimated at training time as
the pose estimate is not involved in the training loss calculation.

Since we assume that a single instance of the object of interest is present in the input image, the set $C$ is assumed to contain only correspondences originating from the single object instance, while being potentially contaminated with outlier correspondences caused by errors in prediction.
The method could be extended to handle multiple instances of the same object, \eg, by using the Progressive-X multi-instance fitting scheme~\cite{barath2019progx}, as in EPOS~\cite{hodan2020epos}.

\section{Experiments}
\label{sec:experiments}

This section analyzes the proposed method for 6DoF object pose estimation and compares its performance with the state-of-the-art methods from the BOP Challenge 2020~\cite{hodan2020bop}. To demonstrate the advantage of predicting dense 3D-3D correspondences, the proposed method is also compared with a baseline that relies on classical 2D-3D correspondences.

\subsection{2D-3D Baseline Method} \label{sec:baseline}

Many state-of-the-art methods for 6DoF object pose estimation build on 2D-3D correspondence estimation~\cite{brachmann2014learning, nocs:wang:cvpr19, zakharov2019dpod, park2019pix2pose, li2019cdpn, hodan2020epos}. While these methods are included in overall evaluation, we also design a directly comparable baseline that uses the same architecture of the feature extractor $F$ and of the subsequent fully-connected network as the proposed method described in Sec.~\ref{sec:method}. However, unlike the fully-connected network $f_\text{NCF}$ which takes a pixel-aligned feature $F(\pi(\mathbf{x}))$ and the depth $\mathbf{x}_z$ of a 3D query point $\mathbf{x}$ and outputs the corresponding 3D coordinates $\mathbf{y}$ and the signed distance $s$, the baseline method relies on a network $f_\text{BL}$ which takes only a pixel-aligned feature $F(\mathbf{p})$ at a pixel $\mathbf{p}$ and outputs the corresponding 3D coordinates $\mathbf{y}$ and the probability $q\in[0, 1]$ that the object is present at $\mathbf{p}$: $f_\text{BL}:\mathbb{R}^K\mapsto\mathbb{R}^3 \times \mathbb{R}$ 
as $f_\text{BL}(F(\mathbf{p}); \boldsymbol{\theta}) = (\mathbf{y}, q)$.
The baseline method is trained by solving the following optimization problem:
\begin{align}\label{eq:blloss}
    &\boldsymbol{\eta}^\star, \boldsymbol{\theta}^\star = \underset{\boldsymbol{\eta,\theta}}{\arg\min} \;
    L_\mathbf{y} + \lambda L_{q} \\
    &= \underset{\boldsymbol{\eta,\theta}}{\arg\min} \;
    \min_{(\bar{R}, \bar{\mathbf{t}}) \in S} \frac{1}{U} \sum_i \bar{q}_i H\left(\bar{R}\mathbf{y}_i + \bar{\mathbf{t}}, \mathbf{x}_i\right) + \lambda \frac{1}{U}\sum_i E\left(q_i, \bar{q}_i\right), \label{eq:loss_baseline}
\end{align}
where $U$ is the number of pixels, $E$ is the softmax cross entropy loss, $\bar{q}$ is given by the ground-truth object mask, and $\bar{\mathbf{y}}$ are the ground-truth 3D coordinates.
At test time, 2D-3D correspondences are established at pixels with $q > 0.5$ and used to fit the object pose with the P\emph{n}P-RANSAC algorithm~\cite{lepetit2009epnp}. In RANSAC, a 2D-3D correspondence $(\mathbf{p}, \mathbf{y})$ is considered an inlier \wrt a pose hypothesis $(R, \mathbf{t})$ if $\|\mathbf{p} - \pi(R\mathbf{y} + \mathbf{t})\|_2$ is below a fixed threshold $\tau_\text{2D}$.

We experiment with two variants of the baseline: ``Baseline-visib'' defines $\bar{q}=1$ for the visible foreground pixels, and ``Baseline-full'' defines $\bar{q}=1$ for all pixels in the object silhouette, even if occluded by other objects.

\subsection{Experimental Setup}

\subsubsection{Evaluation Protocol.} We follow the evaluation protocol of the BOP Challenge 2020~\cite{hodan2020bop}. In short, a method is evaluated on the 6DoF object localization problem, and the error of an estimated pose \wrt the ground-truth pose is calculated by three pose-error functions: Visible Surface Discrepancy (VSD) which treats indistinguishable poses as equivalent by considering only the visible object part, Maximum Symmetry-Aware Surface Distance (MSSD) which considers a set of pre-identified global object symmetries and measures the surface deviation in 3D, and Maximum Symmetry-Aware Projection Distance (MSPD) which considers the object symmetries and measures the perceivable deviation.
An estimated pose is considered correct \wrt a pose-error function~$e$, if $e < \theta_e$, where $e \in \{\text{VSD}, \text{MSSD}, \text{MSPD}\}$ and $\theta_e$ is the threshold of correctness. The fraction of annotated object instances for which a correct pose is estimated is referred to as Recall. The Average Recall \wrt a function~$e$, denoted as $\text{AR}_e$, is defined as the average of the Recall rates calculated for multiple settings of the threshold $\theta_e$ and also for multiple settings of a misalignment tolerance $\tau$ in the case of $\text{VSD}$. The overall accuracy of a method is measured by the Average Recall: $\text{AR} = (\text{AR}_{\text{VSD}} + \text{AR}_{\text{MSSD}} + \text{AR}_{\text{MSPD}}) \, / \, 3$. %

The BOP Challenge 2020 considers the problem of 6DoF localization of a varying number of instances of
a varying number of objects from a single image. To evaluate the proposed method, which was designed to handle a single instance of a single object, we consider only BOP datasets where images show up to one instance of each object. On images that show single instances of multiple objects, we evaluate the proposed method multiple times, each time estimating the pose of a single object instance using the neural networks trained for that object.

\customparagraph{Datasets.} The experiments are conducted on the BOP 2020~\cite{hodan2020bop} version of three datasets: LM~\cite{hinterstoisser2012accv}, LM-O~\cite{brachmann2014learning}, and YCB-V~\cite{xiang2017posecnn}.
The datasets include color 3D object models and RGB-D images of VGA resolution annotated with ground-truth 6DoF object poses (only the RGB channels are used in this work).
LM contains 15 texture-less objects with discriminative color, shape, and
size. Every object is associated with a set of 200 test images, each showing one annotated object instance under significant clutter and no or mild occlusion. LM-O provides ground-truth annotation for instances of eight LM objects in one of the test sets, which introduces challenging test cases with various levels of occlusion.
YCB-V includes 21 objects that are both textured and texture-less, 900 test images showing the objects with occasional occlusions and limited clutter, and 113K real and 80K OpenGL-rendered training images.
Each of these datasets is also associated with 50K physically-based rendered (PBR) images generated by BlenderProc~\cite{denninger2019blenderproc,denninger2020blenderproc} and provided by the BOP organizers. The datasets provide also sets of object symmetry transformations that are used in Eq.~\ref{eq:loss_y} and \ref{eq:loss_baseline}.

\customparagraph{Training.} We report results achieved by the proposed and the baseline methods trained on the synthetic PBR images. On the YCB-V dataset, for which real training images are available, we report also results achieved by the proposed method trained on both real and synthetic PBR images. To reduce the domain gap between the synthetic training and real test images, the training images are augmented by randomly adjusting contrast, brightness, sharpness, and color, as in~\cite{labbe2020cosypose}.
The feature extractor $F$ and networks $f_\text{NCF}$ and $f_\text{BL}$ are initialized with random weights. The networks are optimized by RMSProp~\cite{tieleman2012rmsprop} with the batch size of 4 training images, learning rate of $0.0001$, no learning rate drop, and the balancing weight $\lambda$ set to 1. On LM and LM-O, the optimization is run for 220 epochs. On YCB-V, the optimization is run for 300 epochs on synthetic PBR images, and then for extra 150 epochs on PBR and real images (we report scores before and after the extra epochs).
Special neural networks are trained for each object, while all hyper-parameters are fixed across all objects and datasets.

\customparagraph{Method Parameters.}
The architecture of neural networks is adopted from PIFu~\cite{saito2019pifu}. Specifically, the feature extractor $F$ is a stacked hourglass network with the output stride of $4\,$ and output channel of 256. Networks $f_\text{NCF}$ and $f_\text{BL}$ have four hidden fully-connected layers with 1024, 512, 256 and 128 neurons and with skip connections from $F$.
Unless stated otherwise,
the clamping distance $\delta = 5\,$mm, the inlier threshold $\tau_\text{3D} = 20\,$mm,
the inlier threshold for the baseline method $\tau_\text{2D} = 4\,$px, and the RANSAC-based pose fitting in the proposed and the baseline method is run for a fixed number of 200 iterations.
The sampling step of 3D query points at test time is $10\,$mm (in all three axes) and the near and far planes of the camera frustum, in which the points are sampled, is determined by the range of object distances annotated in the test images (the BOP benchmark explicitly allows using this information at test time).
We converged to these settings by experimenting with different parameter values and optimizing the performance of both the proposed and the baseline method.

In the presented experiments, the signed distance $\psi(\mathbf{y})$
is measured from the query point $\mathbf{x}$ to the closest point on the model surface along the projection ray (\ie, a ray passing through the camera center and $\mathbf{x}$), not to the closest point in 3D as in the conventional SDF~\cite{deepsdf2019park}. However, our additional experiments suggest the two definitions
yield comparable performance.

\begin{table*}[t!]
    \setlength{\tabcolsep}{3pt}
    \scriptsize
	\begin{center}
		\begin{tabularx}{\linewidth}{l Y Y Y Y Y Y Y Y}
		\toprule
		Method &Train & ..type &Test &Refine
        &\text{YCB-V} & ..time &\text{LM-O} &..time \\
\midrule

\Rowcolor{lightgray} NCF (ours)       &rgb &pbr &rgb &-- & \wincat{67.3} & 1.09 & \wincat{63.2} & 4.33 \\
\Rowcolor{lightergray} Baseline-full  &rgb &pbr &rgb &-- & 37.1 & 0.74 & 33.9 & 0.81 \\
\Rowcolor{lightergray} Baseline-visib &rgb &pbr &rgb &-- & 31.9 & 0.71 & 31.6 & 0.79 \\
EPOS~\cite{hodan2020epos}             &rgb &pbr &rgb &-- & 49.9 & 0.76 & 54.7 & 0.47 \\
CDPNv2~\cite{li2019cdpn}              &rgb &pbr &rgb &-- & 39.0 & 0.45 & 62.4 & 0.16 \\

\cmidrule(l{2pt}r{2pt}){6-7}

\Rowcolor{lightgray} NCF (ours)     & rgb &pbr+real &rgb &-- & \wincat{77.5} & 1.09 & \wincat{63.2} & 4.33 \\
leaping 2D-6D~\cite{liu2010leaping} & rgb &pbr+real &rgb &-- & 54.3 & 0.13 & 52.5 & 0.94 \\
CDPNv2~\cite{li2019cdpn}            & rgb &pbr+real &rgb &-- & 53.2 & 0.14 & 62.4 & 0.16 \\
Pix2Pose~\cite{park2019pix2pose}    & rgb &pbr+real &rgb &-- & 45.7 & 1.03 & 36.3 & 1.31 \\

\midrule
CosyPose~\cite{labbe2020cosypose}   & rgb  &pbr+real &rgbd &rc+icp & 86.1 & 2.74 & 71.4 & 8.29 \\
CosyPose~\cite{labbe2020cosypose}   & rgb  &pbr+real &rgb  &rc & 82.1 & 0.24 & 63.3 & 0.55 \\
Pix2Pose~\cite{park2019pix2pose}    & rgb  &pbr+real &rgbd &icp & 78.0 & 2.59 & 58.8 & 5.19 \\
FFB6D~\cite{he2021ffb6d}            & rgbd &pbr      &rgbd &-- & 75.8 & 0.20 & 68.7 & 0.19 \\
K\"onig-Hybrid~\cite{koenig2020hybrid} & rgb &syn+real &rgbd &icp & 70.1 & 2.48 & 63.1 & 0.45 \\
CDPNv2~\cite{li2019cdpn}            & rgb &pbr+real  &rgbd &icp & 61.9 & 0.64 & 63.0 & 0.51 \\
CosyPose~\cite{labbe2020cosypose}   & rgb &pbr &rgb  &rc & 57.4 & 0.34 & 63.3 & 0.55 \\
CDPNv2~\cite{li2019cdpn}            & rgb &pbr &rgbd &icp & 53.2 & 1.03 & 63.0 & 0.51 \\
F{\'e}lix\&Neves~\cite{rodrigues2019deep,raposo2017using} &rgbd &syn+real &rgbd &icp & 51.0 & 54.51 & 39.4 & 61.99 \\
AAE~\cite{sundermeyer2019augmented} &rgb &syn+real &rgbd& icp & 50.5 & 1.58 & 23.7 & 1.20 \\
Vidal~et al.~\cite{vidal2018method} & -- &-- &d &icp & 45.0 & 3.72 & 58.2 & 4.00 \\
CDPN~\cite{li2019cdpn}              & rgb &syn+real &rgb &-- & 42.2 & 0.30 & 37.4 & 0.33 \\
Drost-3D-Only~\cite{drost2010model} & -- & -- & d & icp & 34.4 & 6.27 & 52.7 & 15.95 \\
			\bottomrule
		\end{tabularx}
	\end{center}
    \caption{\label{tab:bop_results} \textbf{Average Recall (AR) scores} on datasets 
	YCB-V and LM-O from BOP 2020~\cite{hodan2020bop}. The 2nd to 5th columns show the training and test setup: image channels used at training (\emph{Train}), type of training images (\emph{Train type}: \emph{pbr} for physically-based rendered images, \emph{syn} for synthetic images which include not only \emph{pbr} images, \emph{real} for real images), image channels used at test (\emph{Test}), and type of iterative pose refinement used at test time (\emph{Refine}: \emph{icp} for a depth-based Iterative Closest Point algorithm, \emph{rc} for a color-based render-and-compare refinement).
	While \emph{pbr} training images are included in both datasets, \emph{real} training images are only in YCB-V -- training setups \emph{pbr} and \emph{pbr+real} are therefore equivalent on LM-O, which leads to several duplicate scores in the table.
	Top scores among methods with the same training and test setup are \wincat{bold}. The time is the average time to estimate poses of all objects in an image [s].
    }
\end{table*}

\subsection{Main Results}

\subsubsection{Accuracy.} Tab.~\ref{tab:bop_results} compares the proposed method (NCF) with participants of the BOP Challenge 2020 and with the baseline method described in Sec.~\ref{sec:baseline}. On the YCB-V dataset, NCF trained on the synthetic PBR images outperforms all competitors which also rely only on RGB images and which do not apply any iterative refinement to the pose estimates.
NCF achieves $17.4\%$ absolute improvement over EPOS~\cite{hodan2020epos} and $28.3\%$ over CDPNv2~\cite{li2019cdpn}, which are trained on the same set of PBR images, and $13.0\%$ and up over~\cite{liu2010leaping,li2019cdpn,park2019pix2pose}, which are trained on PBR and real images. Training on the additional real images improves the AR score of NCF further to $77.5$. Although with smaller margins, NCF outperforms these competitors also on the LM-O dataset. All higher scores reported on the two datasets are achieved by methods that use the depth image or iteratively refine the estimates by ICP or a render-and-compare technique (\emph{c.f.}, \cite{hodan2020bop} for details). On the LM dataset~\cite{hinterstoisser2012accv} (not in Tab.~\ref{tab:bop_results}, see BOP leaderboard~\cite{hodan2020bop}), NCF achieves $81.0$ AR and is close the overall leading method which achives $81.4$ AR and is based on point-pair features~\cite{drost2010model} extracted from depth images.

Tab.~\ref{tab:bop_results} also shows scores of the two variants of the baseline method.
NCF achieves significant improvements over both variants, reaching almost double AR scores. As shown in Tab.~\ref{tab:per_object_results}, NCF outperforms the baseline on all objects from the three datasets. Some of the most noticeable differences are on YCB-V objects 19, 20, and 21. The baseline method struggles due to symmetries of these objects, even though it adopts a very similar symmetry-aware loss as NCF, which performs well on these objects. Qualitative results are in Fig.~\ref{fig:qualitative}.

\customparagraph{Speed.} NCF takes $1.09$ and $4.33\,$s on average to estimate poses of all objects in a test image from YCB-V and LM-O respectively (with a single Nvidia V100 GPU; 3--6 objects are in YCB-V images and 7--8 in LM-O images).
As discussed in Sec.~\ref{sec:ablations}, the processing time can be decreased with sparser query point sampling or with less RANSAC iterations, both yielding only a moderate drop in AR score.
Besides, NCF can be readily used for object tracking, where the exhaustive scanning of the frustum could be replaced by sampling a limited number of query points around the model in the pose estimated in the previous frame.
This would require a lower number of query points and therefore faster processing.

\setlength{\tabcolsep}{3pt}
\begin{table*}[t!]
    \scriptsize
	\begin{center}
    	\begin{tabularx}{\linewidth}{l Y Y Y Y Y Y Y Y Y Y Y Y Y Y Y Y Y Y Y Y Y Y}
    		\toprule
    		
    		\multirow{2}{*}{\vspace{-1.8ex}Method} &
			\multicolumn{8}{c}{LM-O} &
			\multicolumn{14}{c}{LM} \\
			
			\cmidrule(l{5pt}r{-1pt}){2-9}
			\cmidrule(l{5pt}r{-1pt}){10-23}
    		
    		& 1 & 5 & 6 & 8 & 9 & 10' & 11' & 12 & 1 & 2 & 3' & 4 & 5 & 6 & 7 & 8 & 9 & 10' & 11' & 12 & 13 & 14 \\
    		
    		\midrule
    		
    		NCF & \wincat{58} &  \wincat{83} &  \wincat{55} &  \wincat{83} &  \wincat{75} &  \wincat{11} &  \wincat{66} &  \wincat{70} &  \wincat{74} &  \wincat{92} &  \wincat{72} &  \wincat{89} &  \wincat{91} &  \wincat{83} &  \wincat{63} &  \wincat{92} &  \wincat{73} &  \wincat{73} &  \wincat{73} &  \wincat{74} &  \wincat{90} &  \wincat{85} \\
            BL-full & 23 & 53 & 31 & 62 & 38 & \phantom{0}0 & 12 & 40 & 35 & 72 & 45 & 66 & 47 & 47 & 36 & 61 & 45 & \phantom{0}7 & 19 & 49 & 68 & 64 \\
            BL-visib & 25 & 48 & 18 & 46 & 37 & \phantom{0}1 & 27 & 45 & 34 & 69 & 55 & 57 & 48 & 45 & 40 & 54 & 35 & 10 & 29 & 51 & 61 & 47 \\

    		\toprule
    		
    		\multirow{2}{*}{\vspace{-1.8ex}Method} &
			 &
			\multicolumn{20}{c}{YCB-V} \\
			
			\cmidrule(l{5pt}r{-1pt}){2-2}
			\cmidrule(l{5pt}r{-1pt}){3-23}
    		
    		& 15 & 1' & 2 & 3 & 4 & 5 & 6 & 7 & 8 & 9 & 10 & 11 & 12 & 13' & 14 & 15 & 16' & 17 & 18' & 19' & 20' & 21' \\
    		
    		\midrule
    		
    		NCF &  \wincat{83} &  \wincat{67} &  \wincat{81} &  \wincat{83} &  \wincat{57} &  \wincat{77} &  \wincat{72} &  \wincat{76} &  \wincat{75} &  \wincat{51} &  \wincat{85} &  \wincat{84} &  \wincat{72} & \phantom{0}\wincat{8} &  \wincat{61} &  \wincat{84} &  \wincat{36} &  \wincat{63} &  \wincat{41} &  \wincat{60} &  \wincat{62} &  \wincat{49} \\
            BL-full & 57 & 54 & 53 & 66 & 40 & 55 & \phantom{0}6 & 26 & 15 & 31 & 20 & 75 & 36 & \phantom{0}1 & 53 & 60 & \phantom{0}2 & 31 & 18 & \phantom{0}2 & \phantom{0}1 & \phantom{0}0 \\
            BL-visib & 64 & 57 & 40 & 45 & 24 & 52 & 10 & 23 & 32 & 17 & 24 & 55 & 37 & \phantom{0}2 & 40 & 60 & \phantom{0}0 & 25 & 32 & \phantom{0}3 & \phantom{0}0 & \phantom{0}0 \\

    		\bottomrule
        \end{tabularx}
	\end{center}
	\caption{\label{tab:per_object_results} \textbf{Per-object AR scores} on datasets LM-O~\cite{brachmann2014learning}, LM~\cite{hinterstoisser2012accv}, and YCB-V~\cite{xiang2017posecnn}. Objects with symmetries are marked by the prime symbol (').}
\end{table*}

\subsection{Ablation Studies} \label{sec:ablations}

\subsubsection{Performance Under Occlusion.} 
\begin{figure}[t!]
\begin{center}
\begingroup
\renewcommand{\arraystretch}{1.0}
\begin{tabular}{ @{}c@{ } @{}c@{ } }
\includegraphics[width=0.5\linewidth]{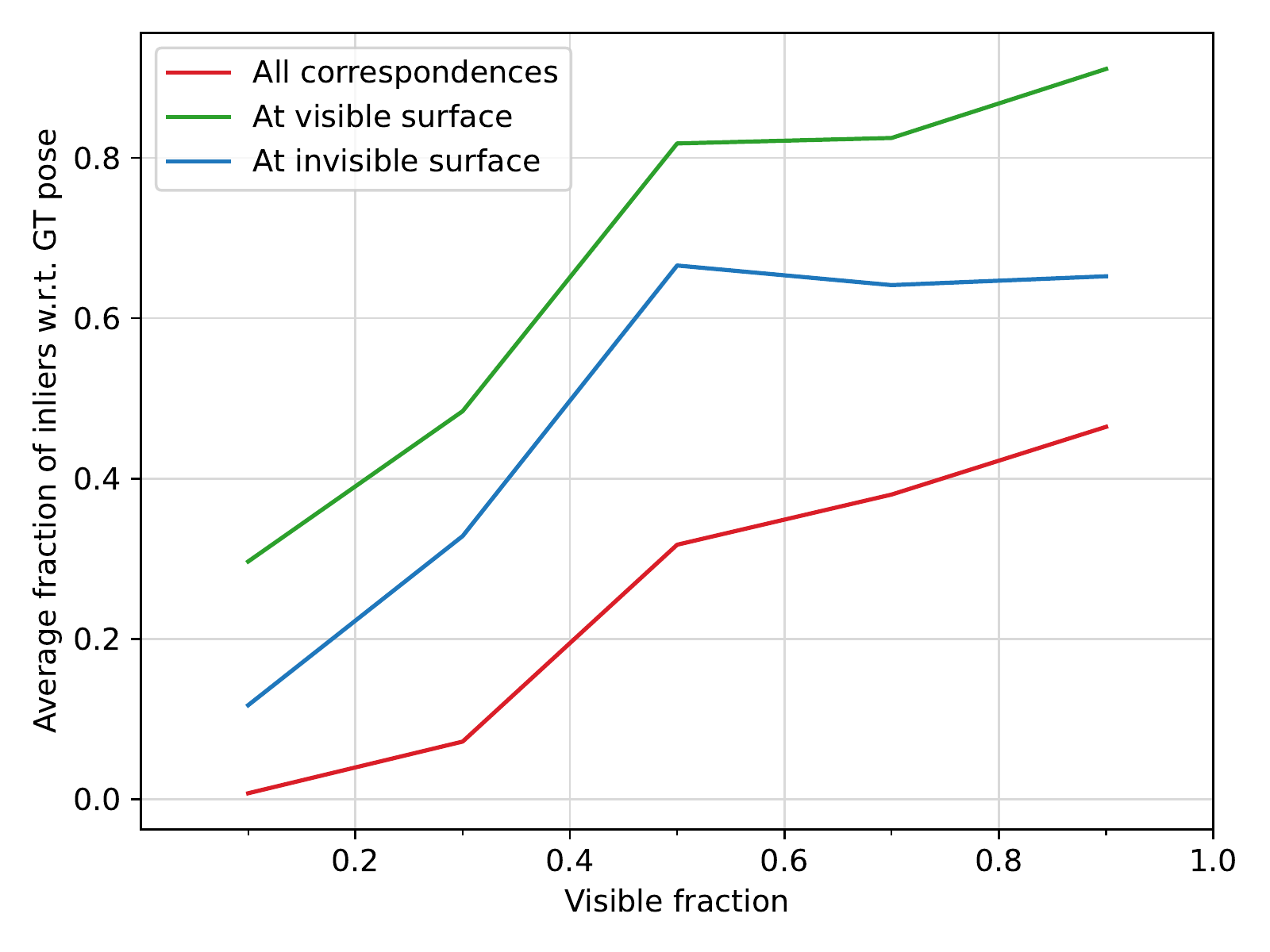} &
\includegraphics[width=0.5\linewidth]{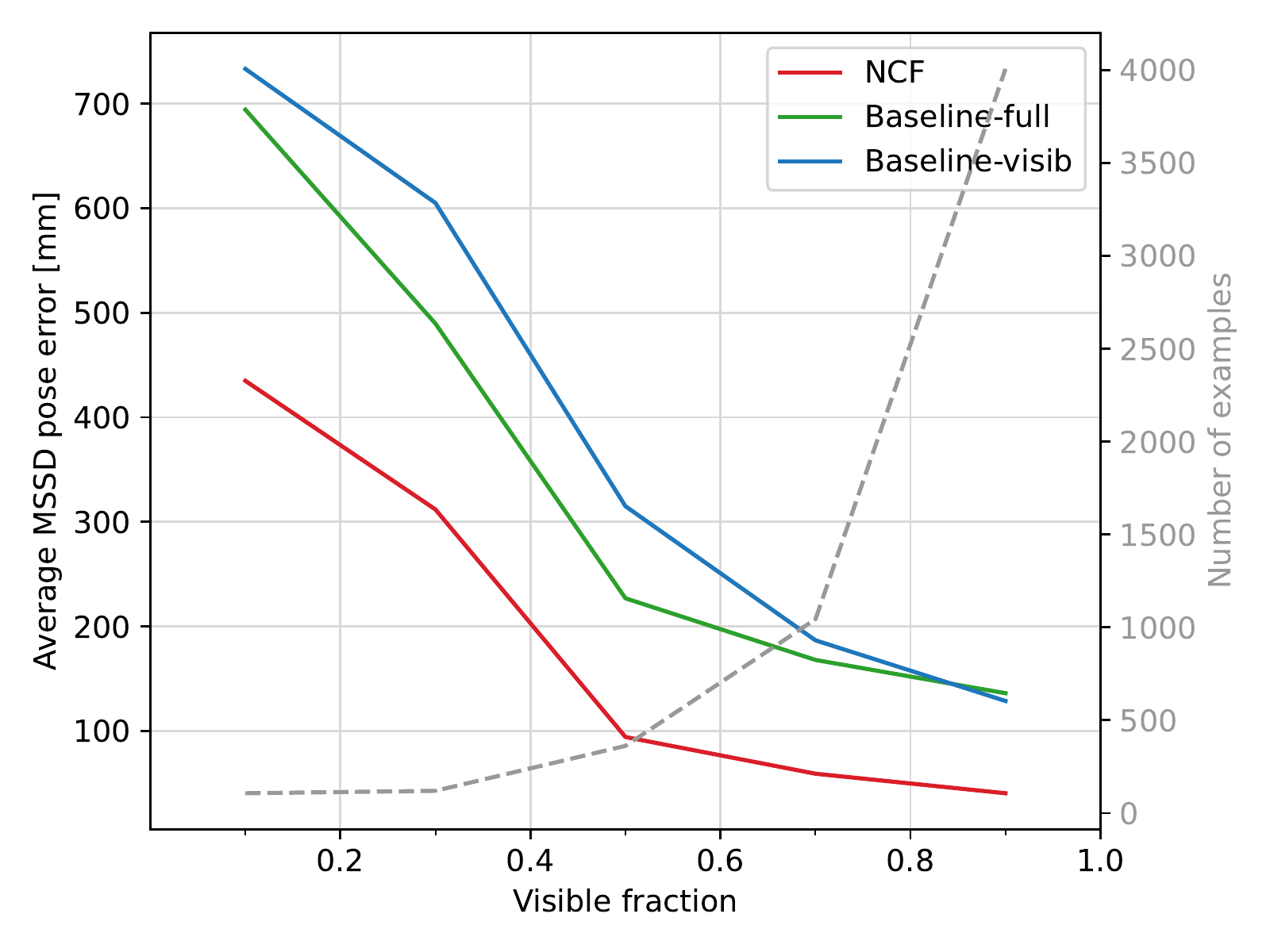} \\
\end{tabular}
\endgroup
\vspace{-1.0ex}
\caption{\label{fig:visibility} \textbf{Performance w.r.t. visible object fraction.} Left: The average fraction of established 3D-3D correspondences that are inliers \wrt the ground-truth pose (\ie, the error of predicted 3D object coordinates is less than a threshold $\tau_\text{3D}=20\,$mm). The set of all correspondences is not the union of correspondences at the visible and invisible surface, hence the red curve is not in between the other two -- see text for details.
Right: The average MSSD error~\cite{hodan2020bop} of object poses estimated by the proposed method (NCF) and the baselines. The average values in both plots are calculated over test examples split into five bins based on the visible fraction of the object silhouette.
}
\vspace{-1.0ex}
\end{center}
\end{figure}

First, we study the impact of different occlusion levels on the quality of predicted 3D-3D correspondences, using the visibility information from~\cite{hodan2020bop}.
This analysis is conducted on datasets YCB-V and LM-O which include partially occluded examples. The quality of correspondences is measured by the fraction of inliers, which is the key metric determining the success of RANSAC~\cite{fischler1981random}. A correspondence $(\mathbf{x}, \mathbf{y})$ is considered an inlier if $\|\bar{R}\mathbf{y} + \bar{\mathbf{t}} - \mathbf{x}\|_2 < \tau_\text{3D} = 20\,$mm, where $\mathbf{x}$ is a 3D query point in the camera coordinates, $\mathbf{y}$ is the predicted 3D point in the model coordinates, and $(\bar{R}, \bar{\mathbf{t}})$ is the ground-truth object pose. Fig.~\ref{fig:visibility} (left) shows the average inlier fraction for test examples split into five bins based on the object visibility.
Already with around $40\%$ visibility (\ie, $60\%$ occlusion), the established correspondences (red curve) include $20\%$ inliers, which is typically sufficient for fitting a good pose with $200$ RANSAC iterations.\footnote{The number of required RANSAC iterations is given by $\log(1 - p) / \log(1 - w^n)$, where $p$ is the desired probability of success, $w$ is the fraction of inliers, and $n$ is the minimal set size~\cite{fischler1981random}. In the discussed case, $p=0.8$ yields $\log(1 - 0.8) / \log(1 - 0.2^3) \approx 200$.}
To separately analyze the quality of correspondences established around the visible and invisible surface, we first select a subset of correspondences established at query points that are in the vicinity of the object surface in the ground-truth pose. This subset is then split into correspondences at the visible surface (green curve in Fig.~\ref{fig:visibility}, left) and at the invisible surface (blue curve).
Although the inlier percentage is higher for correspondences at the visible surface, correspondences at the invisible surface keep up, demonstrating the ability of the proposed method to reason about the whole object.

Next, Fig.~\ref{fig:visibility} (right) shows the impact of occlusion on the average MSSD error~\cite{hodan2020bop} of object poses estimated by the proposed method and the baselines. The proposed method (NCF) clearly outperforms the baselines at all occlusion levels and keeps the average error below $10\,$cm up to around $50\%$ occlusion.

\customparagraph{Density of 3D Query Points.}
The scores discussed so far were obtained with 3D query points sampled with the step of $10\,$mm, \ie, the points are at the centers of $10\times10\times10\,$mm voxels that fill up the camera frustum. On YCB-V, this sampling step yields $230,383$ query points, $0.85\,$s average image processing time and $66.8\,$AR (with 100 RANSAC iterations).
Reducing the step size to $5\,$mm yields $1,852,690$ points and $3.95\,$s, while only slightly improved accuracy of $67.0\,$AR. Enlarging the step size to $20\,$mm yields $28,232$ points, improves the time to $0.63\,$s, and still achieves competitive accuracy of $66.2\,$AR. These results suggest that the method is relatively insensitive to the sampling density.

\customparagraph{Number of Pose Fitting Iterations.} We further investigate the effect of the number of RANSAC iterations on the accuracy and speed. On the YCB-V dataset, reducing the number of iterations from $200$ to $50$ and $10$ decreases the AR score from $67.3$ to $66.7$ and $65.1$, and improves the average processing time from $1.09$ to $0.77$ and $0.68\,$s, respectively. On the other hand, increasing the number of iterations from $200$ to $500$ yields the same AR score and higher average processing time of $1.67\,$s. Note that in the presented experiments we run both Kabsch-RANSAC and P\emph{n}P-RANSAC algorithms for a fixed number of iterations. Further improvements in speed could be achieved by applying an early stopping criterion, which is typically based on the number of inliers \wrt the so-far-the-best pose hypothesis~\cite{fischler1981random,barath2018gcransac}.

\begin{figure}[t!]
	\begin{center}
		\includegraphics[width=1.0\linewidth]{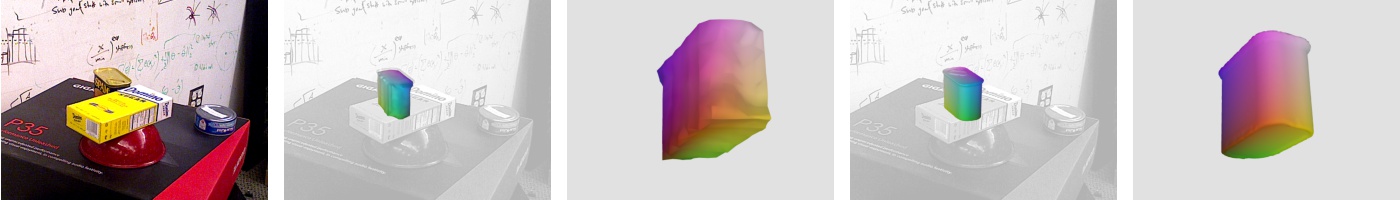} \\ \vspace{0.7ex}
		\includegraphics[width=1.0\linewidth]{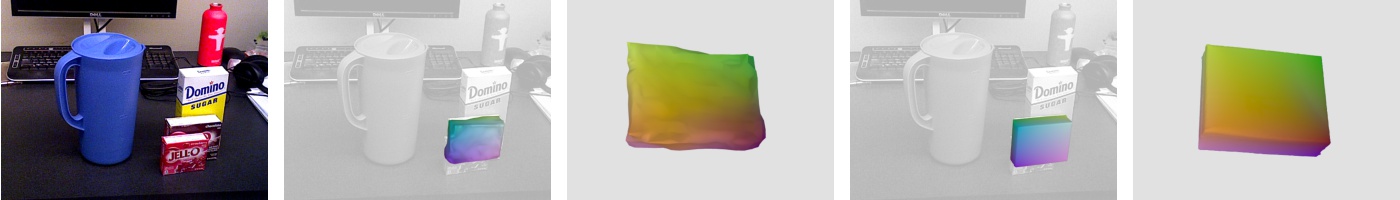} \\ \vspace{0.7ex}
		\includegraphics[width=1.0\linewidth]{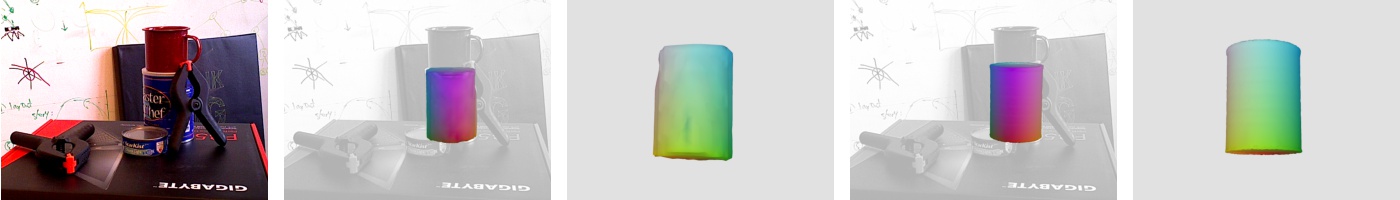} \\ \vspace{0.7ex}
		\includegraphics[width=1.0\linewidth]{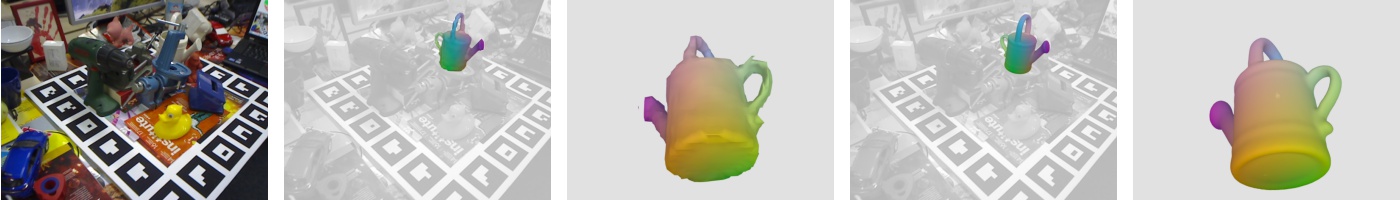} \\ \vspace{0.7ex}
		\includegraphics[width=1.0\linewidth]{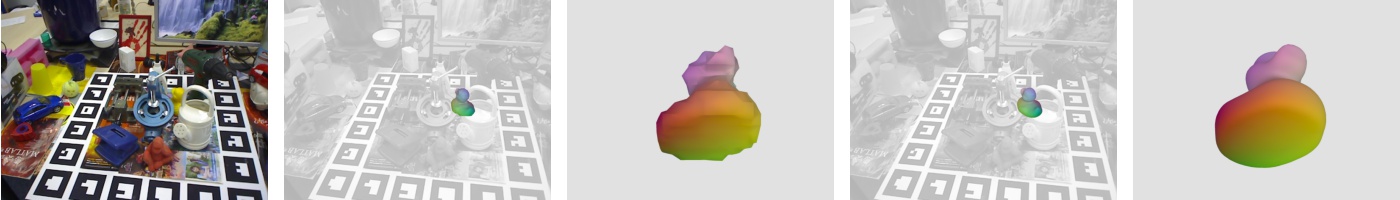} \\ \vspace{0.7ex}
		\includegraphics[width=1.0\linewidth]{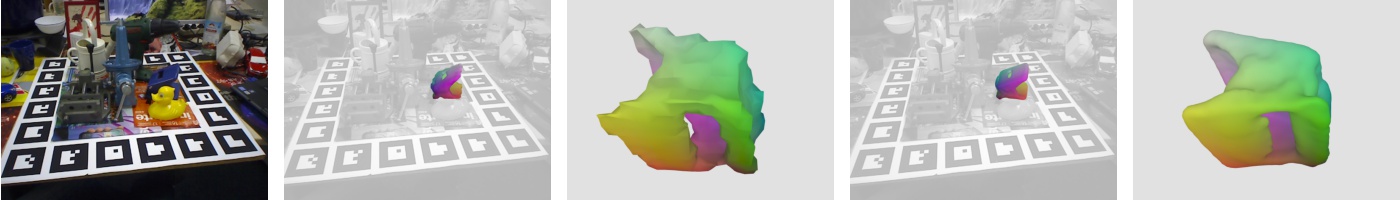} \\
		\setlength{\tabcolsep}{6pt}
        \begin{tabularx}{\linewidth}{Y Y Y Y Y}
            {\small (a)} & {\small (b)} & {\small (c)} & {\small (d)} & {\small (e)}
        \end{tabularx}
		\caption{\label{fig:qualitative} \textbf{Qualitative results on YCB-V and LM-O:} (a) An RGB input. (b) A mesh model reconstructed by Marching Cubes~\cite{lorensen1987marching} from the signed distances predicted at 3D query points in the camera frustum. Mesh vertices are colored with the predicted 3D object coordinates. Note that the mesh is reconstructed only for visualization, not when estimating the object pose. (c)~The reconstructed mesh from a novel view. (d)~GT mesh colored with GT 3D object coordinates. (e) GT mesh in the view from (c).}
	\end{center}
\end{figure}

\section{Conclusion}

We have proposed the first method for 6DoF object pose estimation based on a 3D implicit representation, which we call Neural Correspondence Field (NCF).
The proposed method noticeably outperforms a baseline, which adopts a popular 2D-3D correspondence approach, and also all comparable methods on the YCB-V, LM-O, and LM datasets. Ablation studies and qualitative results demonstrate the ability of NCF to learn and incorporate priors about the whole object surface, which is important for handling challenging cases with occlusion.

\clearpage
\bibliographystyle{splncs04}

\end{document}